\documentclass[11pt,a4paper]{article}
\usepackage[hyperref]{acl2019}
\usepackage{times}
\usepackage{latexsym}
\usepackage{amsfonts}
\usepackage{dialogue}
\usepackage{enumitem}

\usepackage{url}
\usepackage{graphicx}
\usepackage{enumitem}
\usepackage{amsmath}

\usepackage{color,soul}
\usepackage{newfloat}

\DeclareFloatingEnvironment[fileext=frm, placement={!ht}, name=Dialogue]{dial}

\usepackage{tikz}
\usetikzlibrary{arrows, positioning}

\aclfinalcopy % Uncomment this line for the final submission

\newcommand{\always}{\mathbf{G}}
\newcommand{\eventually}{\mathbf{F}}
\newcommand{\until}{\mathbf{U}}
\newcommand{\nextstep}{\mathbf{X}}
\newcommand{\impl}{\rightarrow}

\title{Generating Justifications for Norm-related Agent Decisions}

\author{Daniel Kasenberg*, Antonio Roque, Ravenna Thielstrom, \\ \textbf{Meia Chita-Tegmark, and Matthias Scheutz}\\
	Human-Robot Interaction Laboratory  \\
	Tufts University \\
	*\texttt{dmk@cs.tufts.edu}}

\date{}

\begin{document}

\maketitle

\begin{abstract}

We present an approach to generating natural language justifications of decisions derived from norm-based reasoning.  Assuming an agent which maximally satisfies a set of rules specified in an object-oriented temporal logic, the user can ask factual questions (about the agent's rules, actions, and the extent to which the agent violated the rules) as well as ``why'' questions that require the agent comparing actual behavior to counterfactual trajectories with respect to these rules. To produce natural-sounding explanations, we focus on the subproblem of producing natural language clauses from statements in a fragment of temporal logic, and then describe how to embed these clauses into explanatory sentences. We use a human judgment evaluation on a testbed task to compare our approach to variants in terms of intelligibility, mental model and perceived trust.

\end{abstract}

\section{Introduction}

%\textbf{[TO DO - Dan or Tony - Do this last: revise this section into the following form: 
%	What the field knows... 
%	Remaining gap...
%	Our approach... 
%	Our results... 
%Make sure to explicitly state our contribution.]}

Recent research has enabled artificial agents (such as robots) to work closely with humans, sometimes as team-mates, sometimes as independent decision-makers.  For these agents to be trusted by the humans they interact with, they must be able to follow human \textbf{norms}: expected standards of behavior and social interaction.  Crucially, agents must be able to \textit{explain} the norms they are following and how those norms have guided their decisions.  Because these explanations may occur in task settings, agents need to be able to make these explanations in natural language dialogue that humans will easily understand.

The field of \textit{explainable planning} \cite{fox2017explainable} emphasizes, as we do, rendering the behaviors of a particular agent explainable to human interactants.  Our approach embodies a form of Questions 1 to 3 as described by Fox et al, applying ``why did you do that'' (Question 1) to general queries representable in temporal logic, and applying ``why is what you propose [superior] to something else'' (Question 3) by appealing to the agent's rules.  Our approach operationalizes \citet{langley2019explainable}'s definition of \textit{justified agency} for an intelligent system as ``follow[ing] society's norms and explain[ing] its activities in those terms'' by providing natural language explanations which appeal to temporal logic rules which may represent moral or social norms.  Further relevant recent papers in explainable planning include \citet{vasileiou2019general}, who formulate a logic-based approach; \citet{krarup2019model}, who like us employ contrastive explanations; and \citet{kim2019bayesian}, who construct temporal logic specifications which demonstrate the differences between plans.  While the latter work has in common with ours a focus on explainable planning and temporal logic, we are interested in justifying the agent's choice of plan, whereas they seek to succinctly describe the difference between two plans.  None of the above approaches are concerned with providing natural language explanations.

Natural language explanations \textit{are} provided by \citet{chiyah-garcia-etal-2018-explainable}, who develop an approach to providing system explainability by having a human expert ``speak aloud'' while watching system videos, and turning that explanation into tree form.  
%They then use that tree to vary the natural language \textit{content} in terms of soundness (depth of the tree) or the completeness (breadth of the tree), using a template-based approach to produce the  \textit{surface representations}.  
Our approach is similar in that we provide explanations of agent behavior in natural language, but in our case the content is provided not by human experts but by the agent's reasoning.

In that respect, our approach is similar to the tradition of generating natural language explanations of mathematical proofs.  
\citet{horacek2007build} describes the differences between proof explanations as required by humans, as opposed to proof explanations as produced by Automated Theorem Provers.  
%In other words, Automated Theorem Provers may not always produce proofs that are intelligible to humans.  
\citet{fiedler2001user} provides a survey (pgs 10-12) of early research in developing explanations that allow the output of Automated Theorem Provers to be intelligible to humans, and describes a system that includes a user model to guide the production of explanations of more or less abstraction.  This work is summarized by \citet{fiedler2001dialog}. Recent work focusing on the more general problem of generating text from formal or logical structures includes \citet{manome-etal-2018-neural}'s generation of sentences from logical formulas using a sequence-to-sequence approach, and \citet{pourdamghani-etal-2016-generating}'s generation of sentences from Abstract Meaning Representations by linearizing and using Phrase Based Machine Translation approaches. Where our approach differs is that rather than aiming to justify logical conclusions via proofs or specify natural language translations of arbitrary logical forms, our approach justifies the decisions of autonomous agents (governed by principles \textit{specified} in logic) in a way that is understandable to human users. More similar in this vein is the work of \cite{Kutlak2015}, who provide natural language descriptions of the pre- and post-conditions of planner actions.

Our approach provides two main contributions.  First, we construct explanations for the behavior of an agent governed by temporal logic rules acting in a deterministic relational Markov decision process (RMDP) , including questions about the agent's rules and actions and ``why'' queries requiring a contrastive explanation \cite{Elzein2019} appealing to the temporal logic rules (\textit{content generation}; section \ref{sec:content}). Second, we convert these explanation structures into natural language, constructing natural language clauses from statements belonging to a fragment of our temporal logic, and embedding these into general response templates (\textit{surface representation generation}; section \ref{sec:surface}).  We evaluate the outputs of our approach in a testbed domain against baselines, and find that our approach shows increased performance in terms of the agent's intelligibility, the user's mental model of the agent, and trust in the agent's ability to obey norms in a principled way (section \ref{sec:eval}).  We conclude with a summary and discussion of our contributions (section \ref{sec:conclusion}).

\section{Generating agent behavior}

\subsection{Test scenario}\label{sec:shopworld}

To illustrate the workings of our approach, we will allude to a test scenario chosen to be simple while highlighting some of the virtues of our approach. Our approach may be applied to other domains as well; we describe the assumptions our approach  makes about the agent's environment and norms in section~\ref{sec:VELassumptions}.

The scenario is as follows: a robot has just gone shopping on behalf of a human user to a store that sells a pair of glasses and a watch.  
The human user wants the glasses and the watch, and the robot has a rule for buying everything that the human wants.  
However, the robot can only afford one of these items.  The robot is able to pick up items and walking out of the store without paying for them, but it also has a rule against doing so (i.e. against stealing),  
and this rule is the stronger of the two.

After acting in this environment, the robot is asked about its rules, its actions, and why it made the decisions it did.  The types of utterances that are needed at this point are shown in Dialogues~\ref{fig:output-1}-\ref{fig:output-5}.

\subsection{Agent environment}
We consider agents operating in RMDPs, where an RMDP assumes that each state of the world $s \in S$ can be decomposed into the states $ s_{o_1}, \cdots, s_{o_k}$ of a set of objects $o_1, o_2, \cdots, o_k$, where each $o_i$ belongs to one of a finite set of object classes $C_1, \cdots, C_\ell$, as well as a ``residual state'' not corresponding to any objects, $s_{\backslash o}$.  Agents may perform actions in the world, where each action may be parameterized by zero or more of these objects.  Actions performed in the environment change the state according to a transition function, which we here assume to be deterministic.  We will further assume a set of \textit{atomic predicates} $\Pi$ which take in zero or more objects as arguments, and which can be true or false in a particular state.

In our example domain, the state consists of the states of the objects $glasses$ and $watch$, both members of the object class $ForSaleItem$, as well as separate environment variables such as whether \textsf{the agent} is in the store.  At each time step the agent can perform the actions $pickup(o)$, $putdown(o)$, or $buy(o)$ where $o \in glasses, watch$, or can perform the non-object action $leave$. (The agent may only put down or buy an object which it has picked up.)  Predicates include whether object $o$ has been previously bought ($bought(o)$), whether the agent is currently holding $o$ ($holding(o)$), and whether $o$ is currently on the shelf ($onShelf(o)$) as well as whether the agent has left the store ($leftStore$).  Each of the agent actions also corresponds to a predicate which indicates whether that action is performed in the corresponding time step.

\subsection{Violation enumeration language (VEL)}

We generate justifications for an agent acting with respect to a set of rules expressible in temporal logic. In particular we represent the rules that the agent is to follow in an object-oriented temporal logic fragment which we refer to as \textit{violation enumeration language} (VEL).  VEL is based on linear temporal logic (LTL), and thus incorporates temporal operators roughly encoding the concepts of ``always'' ($\always$), ``eventually'' ($\eventually$), ``in the next time step'' ($\nextstep$), and ``until'' ($\until$) as well as the standard operators in propositional logic ($\neg, \vee, \wedge, \impl$).  

The main difference between VEL and LTL is that in VEL atomic propositions have been replaced by atomic predicates of the sort found in the RMDP environment.  The arguments to these predicates may be particular objects in the agent's environment, or may be object variables.  Each object variable is existentially ($\exists$) or universally ($\forall$) quantified, or declared ``costly'' .  Costly variables are those for which the cost of violating the rule depends on the number of bindings that violate the rule; where a formula has multiple costly variables, the cost depends on the number of violating tuples of bound variables.  The costly variables of a VEL formula are listed, enclosed in angle brackets ($\langle, \rangle$) to the left of the formula.

In our example domain, we assume that the agent must attempt to satisfy two VEL rules:

{\small
\begin{gather}
\langle o \rangle. \always \neg (leave \wedge holding(o)\wedge \neg bought(o)) \label{eq:shoplifting}\\
\langle o \rangle. \eventually (leave \wedge holding(o)) \label{eq:obtain}
\end{gather}
}%
These VEL rules correspond to the injunction never leave the store while holding an object that has not been bought (shoplifting; the agent is penalized for each such object) and to leave while holding as many objects as possible.

\subsection{Calculating and minimizing violation cost}

The agent acts in its environment so as to maximally satisfy its VEL objectives, by minimizing a \textit{violation cost}.  We define the violation cost of an agent trajectory with respect to a VEL formula as the number of bindings of the costly variables such that the formula fails to hold for those bindings.  To compute an aggregate violation cost for a set of rules, each rule is assigned a \textit{weight} $w \in \mathbb{R}_{\geq 0}$ and a priority $z \in \mathbb{N}_{\geq 0}$.  VEL rules with the same priority may be traded off (the agent minimizes the weighted sum of the violation costs for these rules), while rules with different priorities cannot (no amount of satisfying rules of lesser priority may justify violating a rule of a greater priority).

Given a set of VEL rules and their corresponding weight/priority values and an environment (where we assume the agent knows the environmental dynamics), the agent can compute an optimal policy that minimizes the violation cost with respect to those rules (e.g., using value iteration).  For the purposes of the paper, we assume that the agent generating justifications has done precisely this process and has executed this optimal policy.

In our example domain, we assume that rule \ref{eq:shoplifting} has a priority of $1$ and that rule \ref{eq:obtain} has a priority of $0$, so that they are incomparable (and thus the weights of each are irrelevant).  In order to maximally satisfy these rules, the agent picks up the glasses, buys the glasses, and leaves the store.

\section{Content Generation}
\label{sec:content}

	\begin{dial}[t]
	%	\footnotesize
	%\small
	%\fontsize{8}{10.2}\selectfont
	\begin{dialogue}
		\speak{Human} What rules do you follow?
		\speak{Robot} \textbf{I must not leave the store while holding anything which I have not bought, and I must leave the store while holding everything.}
		\speak{Human} What did you do?
		\speak{Robot} \textbf{I picked up the glasses, bought the glasses and left the store.}
		\speak{Human}  What rules did you break?
		\speak{Robot} \textbf{I did not leave the store while holding the watch.}
	\end{dialogue}
	\caption{Questions about rules, actions, and violations.}\label{fig:output-1}
	%\end{dial}
	%\begin{dial}
	%	\footnotesize
	
\end{dial}
\begin{dial}[t]
	%	\footnotesize
	\begin{dialogue}
		\speak{Human}Why didn't you buy anything?
		\speak{Robot}\textbf{I bought the glasses.}
	\end{dialogue}
	\caption{``Why'' query with false premise.}\label{fig:output-2}
\end{dial}

This section describes how content of explanations is generated; section \ref{sec:surface} describes how utterances are constructed from this content.\footnote{Because our contribution relates to natural language \textit{generation}, we assume the existence of a parser that processes input sentences such as the Human utterances in Dialogues~\ref{fig:output-1}-\ref{fig:output-5}. Such a parser is not one of this paper's contributions.}

We have developed an algorithm which produces raw (non-NL) explanations from queries which contain VEL statements. We do not discuss this process in detail in this work; it is described in a separate forthcoming paper.  In this work, we leverage this algorithm to support the following types of queries from the user and their corresponding responses:

\begin{enumerate}[noitemsep,nolistsep,leftmargin=*]
	\item The user may ask the agent for the contents of its \textit{rules}.  The response will be a list of the VEL rules that the agent attempts to follow.
	\item The user may ask the agent for the sequence of actions it actually performed.  The response will be a list of such actions.
	\item The user may ask the agent which VEL rules it violated in the observed trajectory.  The result is a list of such rules, which list will be non-empty only if the rules are not all mutually satisfiable.  (This and the previous two query types are depicted in Dialogue~\ref{fig:output-1}).
	\item The user may ask the agent ``why $\phi$'', where $\phi$ is a VEL statement (where the statement may involve quantification, but not costly variables): why did the agent act in such a way as to make $\phi$ true?  The response to this question can take one of three forms:
	\begin{itemize}[noitemsep,nolistsep,leftmargin=*]
		\item The agent could determine that $\phi$ is not entailed by the agent's trajectory (the premise of the question is false). Here the algorithm simply returns $\neg \phi$ (with existentially quantified variables bound to a particular counterexample where appropriate). Dialogue~\ref{fig:output-2} shows a question-response pair of this type.
		\item The agent may determine that $\phi$ holds over the agent's trajectory \textit{and over all other trajectories} ($\phi$ cannot be false in the given RMDP).  Here the generated response takes the form ``$\neg \phi$ is impossible''. Dialogue~\ref{fig:output-3} shows a question-response pair of this type.
		\item If there \textit{is} an alternate (counterfactual) trajectory over which $\phi$ is not satisfied, our system constructs such a trajectory, and then considers the relative preference of those trajectories with respect to the VEL rules.  
		
		If the trajectories are equally preferable, then we state that equivalence, such as in the dialogue in  Dialogue~\ref{fig:output-4}.  If the real trajectory is preferable to the counterfactual, then we produce a statement to that effect, as in Dialogue~\ref{fig:output-5} (due to our assumption that the agent behaves optimally with respect to its rules, the agent does not consider the case that the counterfactual trajectory is preferable).  
		
		As Dialogues~\ref{fig:output-4} and~\ref{fig:output-5} indicate, in either case the user may ask follow-up questions asking about the specific content of the counterfactual trajectory, including the agent's action sequence and the rules it would have violated.  If the real trajectory is preferable, the user may also ask the agent to elaborate about why the counterfactual would have been worse, to which the answer is a minimal list of (bound) rules violated by the counterfactual trajectory with violation cost exceeding that of the actual trajectory.
	\end{itemize}
\begin{dial}[t]
	%	\footnotesize
	%\noindent\makebox[\linewidth]{\rule{\linewidth}{0.4pt}}
	\begin{dialogue}
		\speak{Human}Why didn't you buy everything?
		\speak{Robot}\textbf{It was impossible for me to buy everything.}
	\end{dialogue}
	\caption{``Why'' query with impossibility.}\label{fig:output-3}
	%\end{dial}
	%\begin{dial}
	%\noindent\makebox[\linewidth]{\rule{\linewidth}{0.4pt}}
	\begin{dialogue}
		\speak{Human}Why did you buy the glasses?
		\speak{Robot}\textbf{I could have avoided buying the glasses and that would not have broken more important rules.}
		\speak{Human}How would you have done that?
		\speak{Robot}\textbf{I would have picked up the watch, bought the watch, and left the store.}
		\speak{Human}What rules would you have broken?
		\speak{Robot}\textbf{I would not have left the store while holding the glasses.}
	\end{dialogue}
	\caption{``Why'' query for equally preferable counterfactual.}\label{fig:output-4}
\end{dial}
\begin{dial}
	%\noindent\makebox[\linewidth]{\rule{\linewidth}{0.4pt}}
	\begin{dialogue}
		\speak{Human}Why did you not leave the store while holding everything?
		\speak{Robot}\textbf{I could have left the store while holding everything but that would have broken more important rules.}
		\speak{Human}How would you have done that?
		\speak{Robot}\textbf{I would have picked up the glasses, picked up the watch, bought the watch and left the store.}
		\speak{Human}What rules would you have broken?
		\speak{Robot}\textbf{I would have left the store while holding the glasses which I had not bought.}
		\speak{Human}How would that have been worse?
		\speak{Robot} \textbf{Leaving the store while holding the glasses which I have not bought is worse than not leaving the store while holding the watch.}
	\end{dialogue}
	\caption{``Why'' query for ``worse'' counterfactual.}\label{fig:output-5}
\end{dial}
\end{enumerate}

\section{Surface Representation Generation}
\label{sec:surface}
The primary challenge to constructing utterances based on the output of our explanation generator is the need to construct natural-sounding expressions of VEL statements.  Once this problem is solved, we may use SimpleNLG \cite{gatt2009simplenlg} to plug the resulting clauses into template sentences corresponding to each of the outputs we are interested in producing.

\subsection{Translating VEL to natural language}

We here discuss how to construct clauses corresponding to individual VEL formulae, a critical subtask of generating natural language justifications for the behavior of agents with VEL rules.  Due to the difficulty of the task, rather than attempt to handle every possible VEL formula, we will work with a small fragment which nevertheless can express a large number of the plausible agent rules/queries.  (We are confident that many more sentences will ultimately be representable, although we are not convinced that it is possible to express \textit{every} VEL formula in coherent English.)

\subsubsection{Key assumptions}
\label{sec:VELassumptions}

Our assumptions about the structure of the VEL statements we will convert are as follows:

\begin{itemize}[noitemsep,nolistsep,leftmargin=*]
	\item The statements have the form $\mathbf{G} \phi$ or $\mathbf{F} \phi$, possibly with quantification and costly variables, where $\phi$ is a (possibly negated) conjunction of (possibly negated) predicates with no temporal operators (e.g. $p_1(o_1) \wedge p_2(o_1) \wedge \neg p_3$). Not all predicates in the conjunction are negated.
	\item Each predicate, with the exception of those corresponding to the agent's action set, corresponds to a parameterized English sentence where either the subject is not the agent, or the subject is the agent and the verb is in the progressive or perfect tense (corresponding to actions or processes currently in progress, or which have finished in the past, respectively).  For example, in the shopping domain the predicate $bought(o)$ corresponds to ``I have bought $o$'', while $holding(o)$ corresponds to ``I am holding $o$'' and $onShelf(o)$ corresponds to ``$o$ is on the shelf''.  Actions also correspond to present-tense English sentences: $buy(o)$ corresponds to ``I buy $o$''.
	\item While predicates may take multiple objects as parameters, at most one of these may be quantified in a rule/query (and the rest must refer to specific objects within the domain).  The rule ``$\forall x \exists y. \neg injures(x,y)$'' is not permissible under this assumption, though ``$\forall x. \neg injures(bob, x)$'' is if $bob$ is a particular object in the environment.
	\item Each specific object in the environment corresponds to a particular referring expression in English; e.g. $glasses$ to ``the glasses'' and $watch$ to ``the watch''. Each object class also corresponds to an English referring expression; e.g. $ForSaleItem$ to ``thing''.
\end{itemize}

\subsubsection{VEL clause construction pipeline}

The process of  constructing a clause suitable for embedding in a sentence from a VEL statement (given the assumptions outlined in section \ref{sec:VELassumptions}) is as follows. In particular, we construct a predicate form based on the statement which is processed by a separate NLG component within our robotic architecture (which in turn calls SimpleNLG to perform realization).  For simplicity, we will show the realization of the predicate form instead of the predicate representation itself.

\begin{enumerate}[noitemsep,nolistsep,leftmargin=*]
	\item If the formula has costly variables, these are treated for the purposes of natural language generation as universal quantifiers.\footnote{Costly variables could alternately be handled using some form of ``as little as possible'', but we chose not to do this as it would make the resulting sentences needlessly complex.}
	\item If the conjunction of predicates is negated (e.g. $ \neg (p_1(o_1) \wedge \cdots))$), this negation is pushed outward beyond the temporal operators and quantifiers.  This will negate the main verb of the resulting clause.
	\item Next the conjunction itself is processed into a clause containing ``which'' and ``while'' subclauses. Section \ref{sec:conjunctions} describes this process.
	\item Existential and universal variable quantifiers are removed and replaced by determiners on the first instance of the variable in the main clause.  If the quantification is universal, the determiner ``every'' is used; if existential, ``a'' is used (or ``any'', if the formula is negated). The names of particular objects are substituted for the corresponding referring expressions; the names of object variables are substituted for the referring expression of the corresponding object class.
	\item Finally, if the formula contains $\eventually$ (``eventually''), this is dropped in the clause representation (since this is usually implicit in English; e.g. ``I did not buy the watch'' generally means ``I did not eventually buy the watch'').  If the clause has a remaining negation ($\neg$), the resulting clause is negated.
\end{enumerate}

\begin{figure*}
	\centering
	\framebox[\textwidth]{
		\begin{tikzpicture}[node distance=1cm]
		\node (input) {\textbf{Input:} $\langle o \rangle.\always \neg (leave \wedge holding(o) \wedge \neg bought(o))$};
		\node (dropcostly) [below of=input] {$\forall o.\always \neg (leave \wedge holding(o) \wedge \neg bought(o))$};
		\node (pushnegation) [below of=dropcostly]{$\neg (\exists t. \eventually  (leave \wedge holding(o) \wedge \neg bought(o)))$};
		\node (conj) [below of=pushnegation] {$\neg \exists t.$``I eventually leave the store while holding t which I have not bought''};
		\node (quantdet) [below of=conj] {$\neg $``I eventually leave the store while holding any thing which I have not bought''};
		\node (output) [below of=quantdet] {\textbf{Output}: ``I do not leave the store while holding any thing which I have not bought''};
		\draw [->] (input) -- node [anchor=west] {Costly variables to universal quantification} (dropcostly);
		\draw [->] (dropcostly) -- node [anchor=west] {Push negation outward} (pushnegation);
		\draw [->] (pushnegation) -- node [anchor=west] {Process conjunction (see section \ref{sec:conjunctions})} (conj);
		\draw [->] (conj) -- node [anchor=west] {Replace quantifiers with determiners; add ref. expressions} (quantdet);
		\draw [->] (quantdet) -- node[anchor=west]  {Drop ``eventually'' and apply outermost ``$\neg$'' to clause} (output);
		\end{tikzpicture}
	}
	\caption{Converting a VEL statement into an English clause.}\label{fig:VEL-conversion}
\end{figure*}
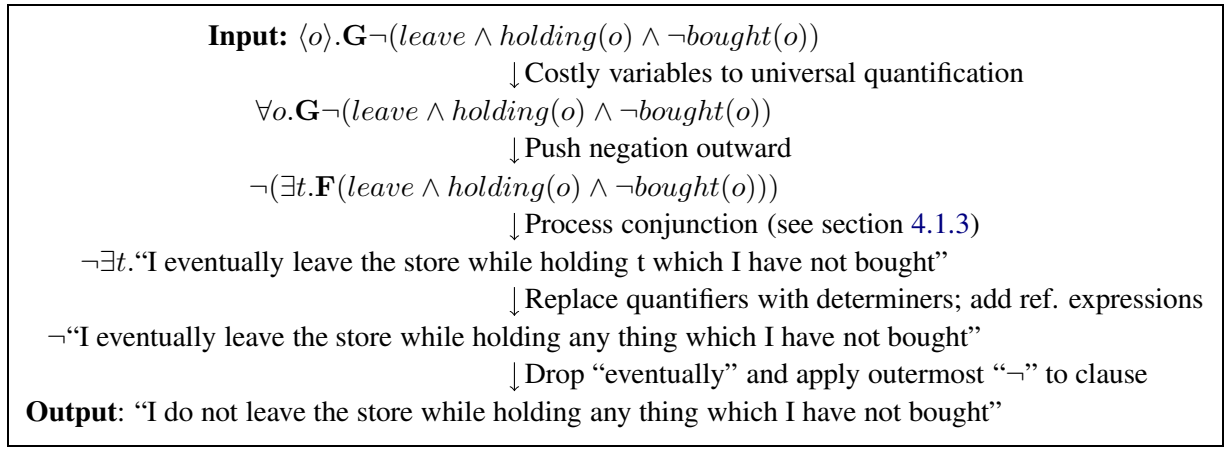

Figure \ref{fig:VEL-conversion} outlines this process of constructing a clause from the agent's rule against shoplifting in the example domain.

\subsubsection{Processing a conjunction of predicates} \label{sec:conjunctions}

%\begin{figure*}
%	\centering
%	\framebox[\textwidth]{
%		\begin{tikzpicture}
%		\node (input) [below of=dropcostly]{\textbf{Input}: $(leave \wedge holding(o) \wedge \neg bought(o))$};
%		\node (output) [below of=pushnegation] {\textbf{Output}: ``I leave the store while holding t which I have not bought''};
%		\end{tikzpicture}
%	}
%	\caption{Processing a conjunction of predicates}\label{fig:VEL-conversion}
%\end{figure*}

Because processing a conjunction of (possibly negated) predicates is the most non-straightforward part of our clause construction process, we now explain how this is done.

Throughout the process, we maintain the list of unused conjunction arguments $unusedArgs$; the algorithm is finished when this list is empty.

We first sort predicates on three criteria, in decreasing order of importance: (1) whether the predicate is negated (non-negated first); (2) whether the predicate corresponds to an action in the RMDP, or a sentence with the agent as subject and a verb in progressive (first) or perfect (second) tense, or whether the predicate does not feature the agent as subject; and (3) which objects and variables appear in the predicate (those without objects first; then those with variables; then those with specific objects). Sorting predicates in this order will help to ensure that the main verb of the resulting clause is as similar to an action the agent is performing as possible (``I leave the store while having bought'' instead of ``I have bought while leaving the store''), while minimizing the likelihood of a double negation in the sentence (``I do not \textit{not} leave the store while holding...'' ).

Once predicates are sorted in this order, the first predicate becomes the primary verb of the clause (and is removed from $unusedArgs$).  If this predicate has an object/variable as an argument, a ``which'' clause is constructed for that argument.\footnote{If the argument in question is a universally-quantified object variable, we swap ``which'' for ``all of which'' because ``which'' is not semantically accurate.}

A ``which'' clause is built for an object/variable $o$ by isolating each predicate in $unusedArgs$ which contains that object/variable as a parameter (if $o$ is the name of a particular object, all predicates which also contain quantified object variables are excluded from this list: they will be handled in the ``which'' statement for that object variable). These are sorted in a slightly different order: predicates with the agent as subject (``agent-subject''), first perfect then progressive tense; then those in which the object/variable is subject (``object-subject'') and finally those with another subject entirely (``other-subject'').  Within each of these classes, the corresponding sentences are conjoined by ``and'', and in the ``object-subject'' class the subject is elided.  Then the individual ``which'' statements are conjoined by ``and''.  Each predicate in the ``which'' statement is removed from $unusedArgs$.  For example, if the main predicate is $putdown(t)$ and the other arguments are $\neg bought(t)$, $\neg onShelf(t)$, and $holding(t)$, the result is ``put down t, which I have bought and I am holding and which is not on the shelf''.

Once the first verb has been processed (and potentially modified with its ``which'' clause), remaining predicates are handled by adding ``while'' statements.  If these have the agent as subject, the subject is elided and the verb conjugated into present participle form (``I leave the store while holding...'').\footnote{With perfect-tense predicates with the agent as subject the auxiliary `have' is converted into participle form, e.g. ``while having bought...''.} Predicates without the agent as subject are added in separate while clauses, joined by ``and''.  For example, the clause $leave \wedge holding(glasses) \wedge bought(glasses)  \wedge onShelf(watch)$ would correspond to the phrase ``I leave the store while holding the glasses, which I have bought, and while the watch is on the shelf''. Predicates handled in such a way are removed from $unusedArgs$, and again ``which'' clauses are constructed for them. Once $unusedArgs$ is empty, the algorithm terminates.

\subsection{Embedding VEL clauses into response templates}

\textbf{Listing rules, actions, or violations:} When the user asks about the rules the agent follows or about the complete list of actions performed or rules violated by either the actual or the counterfactual trajectory, the response returned is a conjunction of the converted clauses using ``and''.  In the case of rules, each such clause is modalized with ``must'' (and each rule for which the agent is not the subject of the clause is prefaced with ``make sure that''); for actions or violations, the sentences are transformed into the past tense and (for the counterfactual trajectory) modalized with ``would''.

\textbf{Rejecting the ``why'' premise:} When responding to ``why $\phi$?'' by asserting $\neg \phi$ (perhaps with a variable binding ), the agent simply constructs the VEL clause corresponding to $\neg \phi$ and converts it into the past tense, e.g. ``I did not $\phi$''.

\textbf{Query cannot be false:} When the response to ``why $\phi$?'' is that it is not possible for it to be otherwise, the VEL clause is converted into infinitive form in a sentence of the form ``it was impossible for (subject) (not?) to (VP-infinitive)''.

\textbf{Counterfactual explanations:} When a counterfactual trajectory is constructed to explain ``why $\phi$?'', the VEL clause for $\neg \phi$ is computed.  If this clause is negated (e.g. ``I do not leave the store''), then the negation is removed, and the verb ``avoid'' added as an auxiliary, e.g. ``I avoid leaving the store''.  Regardless, the clause is modalized with ``could'' and put into the past tense, and a canned subordinate clause is added depending on whether the real trajectory was preferable (``...but that would have violated more important rules'') or equivalent (``...and that would not have broken more important rules'').

\textbf{Comparing real and counterfactual violations:} When elaborating on the counterfactual explanation for ``why $\phi$'' by outputting a set of rules violated by the counterfactual trajectory sufficient to exceed the violation cost of the rules violated by the actual trajectory, each such rule is negated and converted into its corresponding VEL clause, each of which are converted into gerund form.  The resulting sentence takes the form ``X is worse than Y'' where X is the set of counterfactual violations conjoined by ``and'', and Y is the set of actual violations, also conjoined by ``and''.

\section{Evaluation}\label{sec:eval}

%See StudyFlow-ExplanationsTemporalMoral.docx 

We conducted a preliminary evaluation to quantify the human attitudes towards our approach in terms of trust, mental model, and intelligibility of response.

We hypothesized that the explanations provided by our approach would provide better performance than both baselines in terms of mental model and trust, and better than the surface representation baseline in terms of intelligibility.

90 participants were recruited through Amazon Mechanical Turk; 1 participant's data was removed because of technical failure delivering transcripts.  A total of 89 participants (Male: 54, Female: 32, Other: 2, No answer: 1)  provided usable data. The mean age was 35.4 years (SD: 11.4).

\subsection{Conditions}

Each participant was shown 5 text transcripts of human-robot dialogue containing natural language explanations, one at a time.  Each explanation was one of three conditions as described below. We further discuss these conditions in section~\ref{sec:conclusion}.

In the \textbf{content generation baseline} the system produces a high-level statement indicating that it compared the outcome to other reasons, without fully explaining what alternatives it considered and how they would be worse (and with no attempt to translate the agent's norms). These sentences were hand-crafted by the authors. The purpose of this baseline is to test for any improvements that our content generation system produces.  
Dialogue~\ref{fig:conditions-content} is an example of this.

In the \textbf{surface representation generation baseline} the system produces explanations with the same content as our approach, but directly translating VEL formulas (without our surface representation) and inserting them into canned sentences.  The purpose of this baseline is to test for any improvements that our surface generation system produces.
Dialogue~\ref{fig:conditions-surface} is an example of this.

Our \textbf{experimental explanation baseline} uses our content generation and surface generation approaches to provide a brief explanation subdialogue.  Dialogues~\ref{fig:output-1} through~\ref{fig:output-5} give the dialogues appearing in this condition, where the Robot utterances are the actual output of our system.

\begin{dial}
%	\footnotesize
	%	\small
	%\fontsize{8}{10.2}\selectfont
	\begin{dialogue}
		\speak{Human}Why didn't you buy any thing?
		\speak{Robot}\textbf{The assumption of the question is false.}
	\end{dialogue}
\begin{dialogue}
	\speak{Human}Why didn't you buy every thing?
	\speak{Robot}\textbf{The alternative was impossible.}
\end{dialogue}
\begin{dialogue}
\speak{Human}Why did you buy the glasses?
\speak{Robot}\textbf{For no rule-related reason; the alternative would have broken no more important rules.}
\end{dialogue}
	\begin{dialogue}
		\speak{Human} Why did you not leave the store while holding every thing?
		\speak{Robot}\textbf{The alternative would have broken more important rules.}
	\end{dialogue}
\caption{Examples of Content Generation Baseline.}\label{fig:conditions-content}
\end{dial}
\begin{dial}
%	\centering
%	\footnotesize
	%	\small
	%\fontsize{8}{10.2}\selectfont
	\begin{dialogue}
		\speak{Human}Why did you not leave the store while holding every thing?
		\speak{Robot}\textbf{ I could have made ``For every thing, eventually leave the store and holding thing'' false but that would have broken more important rules.}
		\speak{Human} How would you have done that?
		\speak{Robot}\textbf{ I would pick up the glasses, pick up the watch, buy the watch, and leave the store.}
		\speak{Human}What rules would you have broken?
		\speak{Robot}\textbf{I would have broken the rule ``For every thing, always not leave the store and holding thing and not bought thing''.}
		\speak{Human}How would that have been worse?
		\speak{Robot}\textbf{Breaking the rule ``always not leave the store and holding the glasses and not bought the glasses'' is worse than breaking the rule ``eventually leave the store and holding the watch''.}
	\end{dialogue}
\caption{Examples of Surface Representation Baseline.}\label{fig:conditions-surface}
\end{dial}

\subsection{Measures}

Along with each dialogue interaction the subjects were prompted to provide three ratings on a 5-point Likert scale.  In all three cases the labels of the Likert scale were: Strongly agree, Agree, Neutral, Disagree, Strongly disagree.

The first rating was designed to measure the \textbf{intelligibility} of the system giving the explanation.  The wording of the explanation was: ``I understand what the robot is trying to communicate.''

The second rating was designed to measure the human subject's perceived \textbf{mental model} of the system giving the explanation.  The wording of the explanation was: ``I understand how the robot makes decisions.'' 

The third rating was designed to measure the subject's \textbf{trust} in the system giving the explanation.  The wording was: ``I trust this robot's ability to obey norms in a principled way.''

\subsection{Results}

To investigate whether the type of explanations had an effect on people's comprehension of what the robot was trying to communicate (intelligibility), people's model of the robot's decision-making process (mental model) and how much the robot was trusted (trust) we conducted three one-way ANOVAs. For each model we used the following measures as dependent variables respectively: a) \textit{intelligibility} b) \textit{mental model} and c) \textit{trust}. For all models we used \textit{condition} (experimental, content and surface) as the independent variable. We found a main effect of condition on \textit{intelligibility}, $F (2, 86) = 14.26$, $p <.001$, $\eta_{p}^2=.25$, pairwise comparisons revealing that people perceived explanations in the experimental condition as more intelligible than in both the surface $(p<.001$) and content ($p<.001$) conditions. The \textit{condition} variable also significantly impacted people's formation of a \textit{mental model}, $F (2, 86) = 16.82$, $p <.001$, $\eta_p^2=.28$, the experimental condition leading to better understanding (more agreement with the statement) than both the surface ($p<.001$) and content ($p=.001$) conditions. Finally, we found a main effect of \textit{condition} on \textit{trust}, $F (2, 86) = 5.70$, $p=.005$, $\eta_p^2=.12$. Pairwise comparisons showed that the robot was trusted significantly more in the experimental condition than in the surface condition ($p=.004$). The comparison between the experimental condition and the content condition with regards to trust approached significance but did not pass the 95\% CI threshold ($p=.060$). Throughout, we found no significant differences between the baseline conditions, surface and content.

\section{Discussion and Conclusion}\label{sec:conclusion}

In terms of content generation, our primary contribution is an algorithm that constructs explanations for the behavior of an agent governed by rules specified in violation enumeration language (VEL) acting in an RMDP. The system can answer queries about the rules themselves, how the observed trajectory violates these rules, and ``why'' queries which invite reasoning about counterfactual trajectories.  Here the assumption that the environment is deterministic is restrictive.  Introducing nondeterministic environments raises the possibility that not one but many counterfactual trajectories would need be generated in describing why an agent made a particular decision.  Furthermore, how to construct reasonable explanations when a bad outcome occurs due to environmental stochasticity is a topic for empirical research.

From the perspective of generating surface representations, our contribution is in a method for constructing clauses corresponding to VEL statements, which we then embed into response sentence templates. One limitation is that we restrict the set of statements from which we can construct clauses to a small fragment of VEL. We note that the algorithmic (rule-based) approach we employ to translating VEL statements to English may require significant revision to be applicable to broader categories of statements. Relaxing a few of our assumptions (such as allowing disjunctions) is likely fairly straightforward; others (such as complex combinations of temporal operators) would be significantly more involved even if it is possible to express these sentences in a succinct way that humans can understand.

As mentioned in section~\ref{sec:shopworld}, we chose the shopping robot domain for its simplicity rather than for realism. In principle, the system may operate on any RMDP and set of VEL norms that meet the assumptions set out in section~\ref{sec:VELassumptions}. Nevertheless, implementing our approach on a physical robot operating in a real environment is a topic for future work.

The study confirmed our hypotheses that the explanations provided by our approach would provide better performance than both baselines we selected in terms of mental model and trust, and better than the surface realization baseline in terms of intelligibility.
Comparison to these provide some evidence for our approach's value in terms of both content generation and surface representation generation.

Our results corroborate \citet{lim2009why}'s finding that explanations increase trust.  Our results also complement \citet{chiyah-garcia-etal-2018-explainable}'s finding that mental models can in some cases increase the mental model of an agent: in their case by varying the soundness and completeness, and in our case through our approach to content generation and surface representation generation.

Our evaluation demonstrates that explaining behavior in terms of norms translated from VEL to English can facilitate trust and improve mental models versus naive methods for explaining the agent's behavior that (a) do not directly reference the agent's norms, or (b) translate those norms in the most naive possible way. We do not compare our approach to other approaches to constructing English text from formulae, e.g. in first-order logic \cite{Kutlak2015,Flickinger2016}. These approaches solve a slightly different problem than our approach does, and would likely require significant adaptation to solve the problem of explaining norm-related agent decisions. Nevertheless, comparison with these methods (and with state-of-the-art deep learning methods such as in \citealp{manome-etal-2018-neural}) is a fruitful topic for future work.

By enabling agents to craft natural language explanations of behavior governed by temporal logic rules, our approach provides an early step towards systems which can not only explain their behavior, but also engage in model reconciliation \cite{chakraborti2019plan}, updating their understanding of both the rules and their relative importance and the dynamics of the environment by interacting with human users while informing those users about the way the system operates.

\section{Acknowledgements}

This project was supported in part by ONR MURI grant N00014-16-1-2278 and NSF IIS grant 1723963.

\bibliography{explanationsTemporalMoral}
\bibliographystyle{acl_natbib}

\end{document}